\pdfoutput=1

\documentclass[11pt]{article}

\usepackage{emnlp2021}

\usepackage{times}
\usepackage{latexsym}

\usepackage[T1]{fontenc}
\usepackage[utf8]{inputenc}
\usepackage{CJKutf8}

\usepackage{graphicx}
\usepackage{grffile}
\usepackage{multirow}
\usepackage{xcolor,colortbl}
\usepackage{amsmath}
\usepackage{amssymb}
\usepackage{makecell}
\usepackage[super]{nth}
\usepackage{arydshln}

\usepackage{algorithm}
\usepackage[noend]{algpseudocode}
\usepackage{regexpatch}
\usepackage{subcaption}
\usepackage{microtype}

\newcommand{\src}{\ensuremath{\mathbf{f}}} 
\newcommand{\trg}{\ensuremath{\mathbf{e}}} 

\title{One Source, Two Targets: Challenges and Rewards of Dual Decoding}

\author{Jitao Xu \\
  Univ. Paris-Saclay, \\
  \& CNRS, LISN \\
  Orsay, France \\
  \texttt{jitao.xu@limsi.fr} \\\And
  François Yvon \\
  Univ. Paris-Saclay, \\
  \& CNRS, LISN \\
  Orsay, France \\
  \texttt{francois.yvon@limsi.fr} \\}

\begin{document}
\maketitle
\begin{abstract}
Machine translation is generally understood as generating one target text from an input source document. In this paper, we consider a stronger requirement: to jointly generate two texts so that each output side effectively depends on the other. As we discuss, such a device serves several practical purposes, from multi-target machine translation to the generation of controlled variations of the target text. We present an analysis of possible implementations of dual decoding, and experiment with four applications. Viewing the problem from multiple angles allows us to better highlight the challenges of dual decoding and to also thoroughly analyze the benefits of generating matched, rather than independent, translations.
\end{abstract}

\section{Introduction \label{sec:introduction}}

Neural Machine Translation (NMT) is progressing at a rapid pace. Since the introduction of the first encoder-decoder architecture \cite{Sutskever14sequence,Cho14properties}, then completed with an attention mechanism \cite{Bahdanau15neural,Vaswani17attention}, the performance of NMT systems is now good enough for a growing number of services, both for the general public and the translation industry. Not only are neural translation systems better, they are also more versatile and have been extended in many ways to meet new application demands. This is notably the case with multilingual extensions \citep{Firat16multiway,Ha16towards,Johnson17googles}, which aim to develop systems capable of processing multiple translation directions with one single model.

Another common situation for MT applications is the multi-source / multi-target scenario, where source documents in language $S_l$ need to be published in several target languages ${T^1_l}, {T^2_l}, \dots.$ This is, for instance, what happens in multilingual institutions, or with crowdsourced translations of TV shows. The multi-source way \cite{Och01multisource,Schwartz08multisource,Zoph16multi-source} to handle this generates a first translation into target language ${T^1_l}$, which, once revised, can be used in conjunction with the original source to generate the translation into language ${T^2_l}$. The expected benefit of this approach is to facilitate word disambiguation.

An alternative, that we thoroughly explore here, is to \emph{simultaneously generate} translations in ${T^1_l}$ and ${T^2_l}$, an approach termed \emph{multi-target translation} by \citet{Neubig15multi}. While the same goal is achieved with a multilingual system translating independently in ${T^1_l}$ and ${T^2_l}$, several pay-offs are expected from a joint decoding: (a) improved disambiguation capacities (as for multi-source systems); (b) a better collaboration between the stronger and the weaker decoders; (c) more consistent translations in ${T^1_l}$ and ${T^2_l}$ than if they were performed independently. As it turns out, a dual decoder computing joint translations can be used for several other purposes, that we also consider: to simultaneously \emph{decode in two directions}, providing a new implementation of the idea of \citet{Watanabe02bidirectional,Finch09bidirectional}; to \emph{disentangle mixed language} (code-switched) texts into their two languages \citep{Xu21traducir}; finally, to generate \emph{coherent translation alternatives}, an idea we use to compute polite and impolite variants of the same input \citep{Sennrich16controlling}.
\begin{CJK*}{UTF8}{gbsn}
\begin{table}
  \centering\footnotesize
  \begin{tabular}{cl}
    \src & I could do that again if you want . \\ \hline
    \trg$_1$ & 只要\ 你\ 愿意\ 我\ 可以\ 重复\ 一遍\ 。\\
    \trg$_2$ & もう\ 一\ 回\ やり\ ましょ\ う\ か\ \\ \hline
    \trg$_1$ & Je peux le refaire si vous le voulez . \\
    \trg$_2$ & . voulez le vous si refaire le peux Je \\ \hline
    \trg$_1$ & Ich kann das noch mal machen , wenn Sie wollen . \\
    \trg$_2$ & Ich kann das noch mal machen , wenn du willst . \\ \hline    
  \end{tabular}
  \caption{Instances of Dual Decoding: multi-target translation (\textsection~\ref{sec:multi-target}), bi-directional decoding (\textsection~\ref{sec:bi-directional}), variant generation (\textsection~\ref{sec:variants}).}
  \label{tab:examples}
\end{table}
\end{CJK*}
Considering multiple applications allows us to assess the challenges and rewards of dual decoding under various angles and to better evaluate the actual agreement between the two decoders' outputs. Our main contributions are the following: (i) a comparative study of architectures for dual decoding (\textsection~\ref{sec:architecture}); (ii) four short experimental studies where we use these architectures to simultaneously generate several outputs from one input (\textsection~\ref{sec:multi-target}--\textsection~\ref{sec:variants}); (iii) practical remedies to the shortage of multi-parallel corpora that are necessary to implement multi-target decoding; (iv) concrete solutions to mitigate exposure bias between two decoders; (v) quantitative evaluations of the increased consistency incurred by a tight interaction between decoders. An additional empirical finding that is of practical value is the benefits of exploiting multi-parallel corpora to fine-tune multilingual systems.

\section{Architectures for Dual Decoding\label{sec:architecture}}

\subsection{Model and Notations \label{ssec:model}}
In our setting, we consider the simultaneous translation of sentence $\src$ in source language $S_l$ into two target sentences $\trg^1$ and $\trg^2$ in languages\footnote{In our applications, these do not always correspond to actual natural languages. We keep the term for expository purposes.} ${T^{1}_l}$ and ${T^{2}_l}$. In this situation, various modeling choices can be entertained \citep{Le20dual}: 
\begin{align}
  P(\trg^1,\trg^2|\src) &= \prod_{t=1}^{T} P(\trg^{1}_{t}, \trg^{2}_{t}|\src, \trg^{1}_{<t}, \trg^{2}_{<t})  \label{eq:depgen}\\ 
  P(\trg^1,\trg^2|\src) &= \prod_{t=1}^{T} P(\trg^{1}_{t}|\src, \trg^{1}_{<t}, \trg^{2}_{<t}) \times \nonumber \\
                        & \quad\quad P(\trg^{2}_{t}|\src, \trg^{1}_{<t}, \trg^{2}_{<t}) \label{eq:jointgen} \\ %
  P(\trg^1,\trg^2|\src)  &= \prod_{t=1}^{T} P(\trg^{1}_{t}|\src, \trg^{1}_{<t}) P(\trg^{2}_{t}|\src, \trg^{2}_{<t}), \label{eq:indepgen}
\end{align}
where $T= max(|\trg^1|,|\trg^2|)$, and we use placeholders whenever necessary. The factorization in Equation~\eqref{eq:depgen} implies a joint event space for the two languages and a computational cost we deemed unreasonable. We instead resorted to the second (\emph{dual}) formulation, that we contrasted with the third one (\emph{independent} generation) in our experiments. Note that thanks to asynchronous decoding, introduced in Section~\ref{ssec:asynchronous}, we are also in a position to simulate other dependency patterns, where each symbol $\trg^2_t$ is generated conditioned on $\trg^{1}_{<t+k}, \trg^{2}_{<t}$, thus reproducing the \emph{chained} model of \citet{Le20dual}.

\subsection{Attention Mechanism\label{ssec:attention}}
Our dual decoder model implements the encoder-decoder architecture of the Transformer model of \citep{Vaswani17attention}. In this model, the input to each attention head consists of queries $\mathbf{Q}$, key-value pairs $\mathbf{K}$ and $\mathbf{V}$. Each head maps a query and a set of key-value pairs to an output, computed as a weighted sum of the values, where weights are based on a compatibility assessment between query and keys, according to (in matrix notations):
\begin{equation}
  \operatorname{Attention}(\mathbf{Q}, \mathbf{K}, \mathbf{V}) = \operatorname{softmax}(\frac{\mathbf{Q}\mathbf{K}^T}{\sqrt{d_k}})\mathbf{V},
  \label{eq:attn}
\end{equation}
with $d_k$ the shared dimension of queries and keys. Note that $\mathbf{Q}$, $\mathbf{K}$ and $\mathbf{V}$ for each head are obtained by linearly transforming the hidden states from previous layer with different projection matrices. 

\subsection{Proposal for a Dual Decoder\label{ssec:dual}}

\begin{figure}[htbp]
  \center
  \includegraphics[width=0.9\columnwidth]{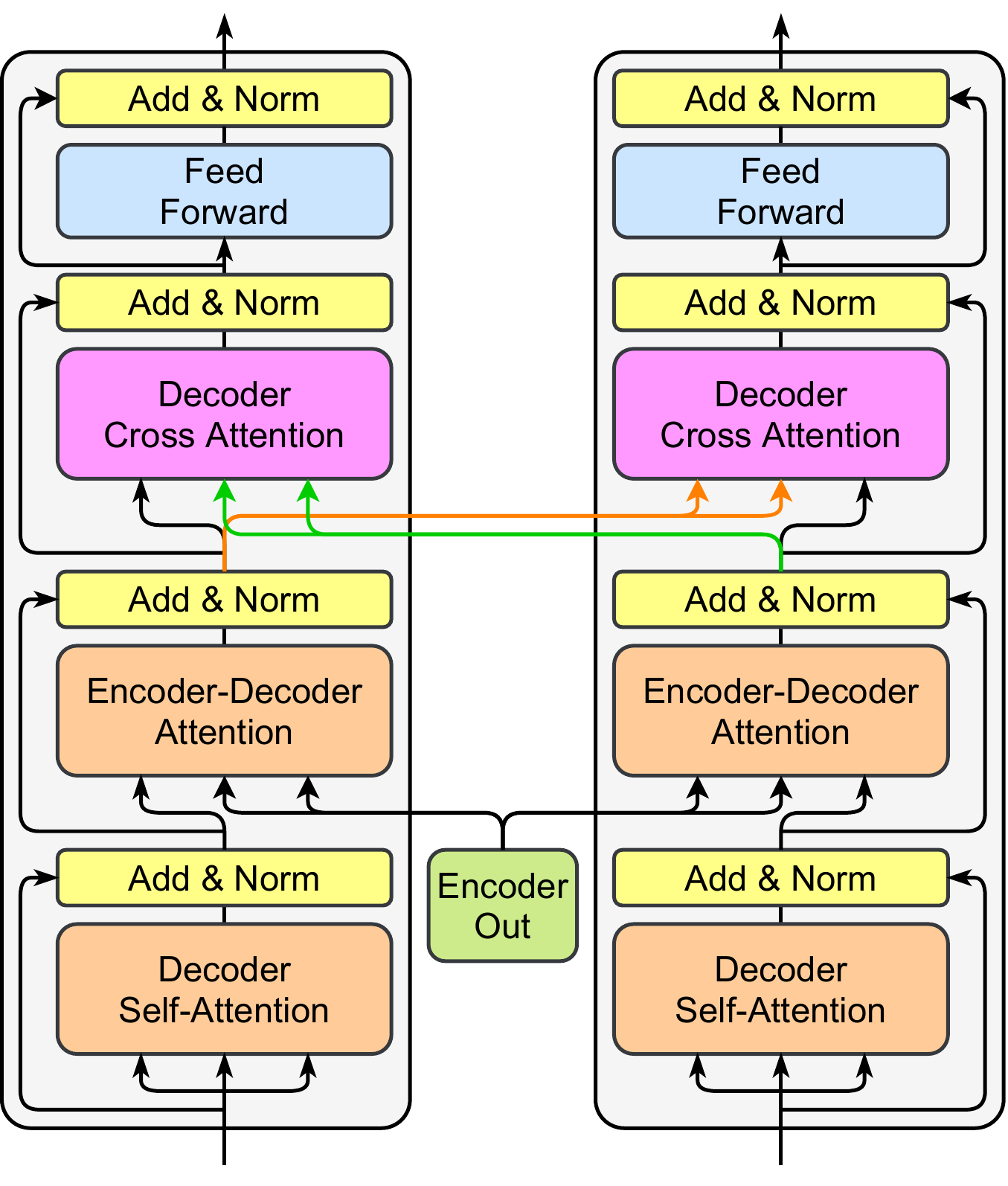}
  \caption{A graphical view of the dual decoder.\label{fig:model}}
\end{figure}

We chose to implement Equation~\eqref{eq:jointgen}  with a synchronous coupling of two decoders sharing the same encoder. An alternative would be to have the two decoders share the bottom layers and only specialize at the upper layer(s) for one specific language: we did not explore this idea further, as it seemed less appropriate for the variety of applications considered. Figure~\ref{fig:model} illustrates this design. Compared to a standard Transformer, we add a cross attention layer in each decoder block to capture the interaction between the two decoders. Denoting the output hidden states of the previous layer for each decoder as $H^1_l$ and $H^2_l$, the decoder cross-attention is computed as:\footnote{For simplicity, we omit the other sub-layers (self attention, encoder-decoder cross attention, feed forward and layer normalization).}
\begin{equation}
  \begin{split}
  H^1_{l+1} &= \operatorname{Attention}(H^1_l, H^2_l, H^2_l) \\
  H^2_{l+1} &= \operatorname{Attention}(H^2_l, H^1_l, H^1_l),
  \end{split}
  \label{eq:mutual}
\end{equation}
where $\operatorname{Attention}$ is defined in Equation~\eqref{eq:attn}. The two decoders are thus fully synchronous as each requires the hidden states of the other in each block to compute its own hidden states. The decoder cross-attention can be inserted before or after the encoder-decoder attention. Preliminary experiments with these variants have shown that they were performing similarly. We thus only report results obtained with the decoder cross-attention as the last attention component of a block (see Figure~\ref{fig:model}).

\subsection{Synchronous Beam Search\label{ssec:beam-search}}

\subsubsection{Full Synchronous Mode}
Our decoding algorithm uses a dual beam search.
Assuming each decoder uses its own beam of size $k$, the cross-attention between decoders can be designed and implemented in multiple ways: for instance, one could have each hypothesis in decoder~1 attend to any hypotheses in decoder~2, which would however create an exponential blow-up of the search space. Following \citet{Zhou19synchronous}, we only compute the attention between 1-best candidates of each decoder, 2-best candidates in each decoder, etc. This heuristic ensures that the number of candidates in each decoder beam remains fixed.
There is however an added complexity, due to the fact that the ranking of hypotheses in each decoder beam evolves over time: the best hypothesis in decoder~1 at time $t$ may no longer be the best at time $t+1$.
Preserving the consistency of the decoder states therefore implies to recompute the entire prefix representation for each hypothesis and each decoder at each time step, thus creating a significant computing overhead.\footnote{In our experiments, the dual decoder was about twice as slow as the two independent ones.}

We also explored other implementations where each candidate prefix in one beam always attends to the best candidate in the other beam or attends to the average of all candidates. These variants ended up delivering very similar results, as also found in \citep{Zhou19synchronous}. For simplicity reasons, we use the first scheme in all experiments.

\subsubsection{Relaxing Synchronicity \label{ssec:asynchronous}}

Simultaneously generating symbols in two languages is a very strong requirement, and may not bring out all the benefits of dual decoding, especially when the two target languages have different word orders. We relax this assumption by allowing one decoder to start generating symbols before the other: this is implemented by having the delayed decoder generate dummy symbols for a fixed number of steps before generating meaningful words, a strategy akin to the wait-k approach in spoken language translation \citep{Elbayad20waitk}.

A more extreme case of delayed processing is when one decoder can access a \emph{complete translation} in the other language. In our implementation, this is simulated with partial forced decoding, where one translation is predefined, while the other is computed. We  explored this in two settings: (a) within a \emph{two-pass, sequential procedure}, where the output of step~1 for decoder~1 is fully known and fixed when computing the second output of decoder~2; (b) using a \emph{reference translation} in one of the decoder, implementing a controlled decoding where the output in ${T^2_l}$ not only translates the source, but does so in a way that is consistent with the reference translation in ${T^1_l}$. These strategies are used in Sections~\ref{sec:multi-target} and \ref{sec:variants}.

\subsection{Training and Fine-tuning\label{ssec:train}}
Training this model requires triplets of instances comprising one source and two targets. Given a set of such examples $D=\{(\src, \trg^1, \trg^2)_i, i=1 \dots N\}$, the training maximizes the combined log-likelihood for the two target sequences:
\begin{equation}
  \begin{split}
  L(\theta) = \sum_{D} (\sum_{t=1}^{|\trg^1|}\log P(\trg_t^1 | \trg^1_{<t}, \trg^2_{<t}, \src, \theta) \\
  + \sum_{t=1}^{|\trg^2|}\log P(\trg^2_t|\trg^2_{<t}, \trg^1_{<t}, \src, \theta)),
  \end{split}
\end{equation}
where $\theta$ represents the set of parameters.

As multi-parallel corpora are not as common as bilingual ones, we also considered a two-step procedure which combines bilingual and trilingual data. In a first step, we train a standard multilingual model (one monolingual encoder, one bilingual decoder), where tags are used to select the target language \citep{Johnson17googles}. This only requires bilingual data $\{(\src, \trg^1)_i, i=1 \dots N^1\}$ and $\{(\src', \trg^2)_j, j=1 \dots{} N^2\}$. We then initialize the dual decoder model with pre-trained parameters and fine-tune with the trilingual dataset. Both decoders thus start with the same pre-trained decoder. The decoder cross-attention matrices cannot benefit from pre-training and are initialized randomly. During fine-tuning, tags are no longer necessary as both target translations are required.

\section{Multi-target Machine Translation \label{sec:multi-target}}

\subsection{Data\label{ssec:data}}
We first evaluate our dual decoder model on the multi-target MT task for three directions: English to German/French (En$\to$De/Fr), German to English/French (De$\to$En/Fr) and English to Chinese/Japanese (En$\to$Zh/Ja). Similarly to \citep{Wang19synchronously,He21synchronous}, we use the IWSLT17 dataset\footnote{\url{https://wit3.fbk.eu/2017-01-c}} \citep{Cettolo12wit3} as our main test bed.\footnote{Subject to some filtering to obtain a fully multi-parallel set of sentences.}
Pre-training experiments additionally use WMT20 De-En, De-Fr, En-Zh, En-Ja and WMT14 En-Fr bilingual datasets.\footnote{See \url{http://statmt.org}.} We use the IWSLT \texttt{tst2012} and \texttt{tst2013} as development sets and test our model on \texttt{tst2014}.\footnote{For comparison with \citep{He21synchronous}, we also report the results on \texttt{tst2015} in Appendix \ref{sec:res-appendix}.}
Table~\ref{tab:data} summarizes the main statistics for trilingual training and test data.

\begin{table}[!ht]
  \center
  \scalebox{0.85}{
  \begin{tabular}{l|cc|c}
  \hline
       & Original De & Original Fr & 3-way \\
  \hline
  Train &    209522  &   236653    &  205397   \\
  Dev   &    2693    &    3083     &  2468     \\
  tst2014 &    1305  &    1306     &  1168     \\
  \hline
       & Original Zh & Original Ja & 3-way \\
  \hline
  Train    &     235078  &    226834   & 213090    \\
  Dev      &     3064    &     3024    & 2837      \\
  tst2014  &     1297    &     1285    & 1214      \\
  \hline
  \end{tabular}
  }
  \caption{Number of lines in the trilingual IWSLT data. English is used to identify trilingual sentences and is therefore not shown in this table.\label{tab:data}}
\end{table}

For WMT data, we discard sentence pairs with invalid language tag as computed by \texttt{fasttext} language identification model\footnote{\url{https://dl.fbaipublicfiles.com/fasttext/supervised-models/lid.176.bin}} \citep{Bojanowski17enriching}. 
We tokenize all English, German and French data using Moses tokenizer.\footnote{\url{https://github.com/moses-smt/mosesdecoder}} Chinese and Japanese sentences are segmented using \texttt{jieba}\footnote{\url{https://github.com/fxsjy/jieba}} and \texttt{mecab}\footnote{\url{https://taku910.github.io/mecab/}} respectively. For En$\to$De/Fr and De$\to$En/Fr, we use a shared source-target vocabulary built with 40K Byte Pair Encoding (BPE) units \citep{Sennrich16BPE} learned on WMT data with \texttt{subword-nmt}.\footnote{\url{https://github.com/rsennrich/subword-nmt}} For En$\to$Zh/Ja, we build a 32K BPE model for En and a joint 32K BPE for Zh and Ja, both learned on the WMT data.\footnote{Full experimental details are in Appendix \ref{sec:detail-mtmt}.}

\subsection{Experimental Settings\label{ssec:settings}}

We implement the dual decoder model using \texttt{fairseq}\footnote{\url{https://github.com/pytorch/fairseq}} \citep{Ott19fairseq},\footnote{Our implementation is open-sourced at \url{https://github.com/jitao-xu/dual-decoding}.} with a hidden size of $512$ and a feedforward size of $2048$. We optimize with Adam, set up with a maximum learning rate of $0.0007$ and an inverse square root decay schedule, as well as $4000$ warmup steps. For fine-tuning models, we use Adam with a fixed learning rate of $8\mathrm{e}{-5}$. For standard Transformer models, we share the decoder input and output matrices, while for dual decoder models, we share all four input and output decoder matrices \citep{Press17using,Inan17tying}. All models are trained with mixed precision and a batch size of $8192$ tokens on 4 V100 GPUs. Pre-training last for $300$k iterations, while all other models are trained until no improvement is found for $4$ consecutive checkpoints on the development set. Performance is computed with SacreBLEU \citep{Post18sacrebleu}.

We call the dual decoder models \texttt{dual}. To study the effectiveness of dual decoding, we also train a simplified multi-task model \citep{Dong15multitask}, implementing the independent model of Equation~\eqref{eq:indepgen} without decoder cross-attention. For \texttt{indep},  the only interaction between outputs is thus a shared loss computed on multi-parallel data. Baseline Transformer models trained separately on each language pair are denoted by \texttt{base}.

\begin{table*}[!ht]
  \center
  \scalebox{0.8}{
  \begin{tabular}{l|ccc|ccc|ccc|l|l}
  \hline
  Model  & De-En & De-Fr & SIM & En-De & En-Fr & SIM & En-Zh & En-Ja & SIM & Avg BLEU & Avg SIM\\
  \hline
  \texttt{base}      & 32.6 & 24.8 & 89.28 & 28.1 & 38.8 & 91.34 & 22.2 & 13.6 & 81.97 & 25.5        & 87.53 \\
  \texttt{multi} & 31.1 & 24.4 & 91.56 & 25.9 & 37.9 & 91.87 & 25.3 & 10.4 & 83.71 & 25.8 (+0.3) & 89.05 (+1.52) \\
  \hline
  \texttt{indep}  & 33.8 & 25.4 & 90.04 & 29.1 & 39.8 & 91.63 & 22.6 & 14.8 & 83.16 & 26.3 (+0.8) & 88.28 (+0.75) \\
  \texttt{dual}   & 31.8 & 22.4 & 90.60 & 28.5 & 38.8 & 91.45 & 22.8 & 15.3 & 84.09 & 25.6 (+0.1) & 88.71 (+1.18) \\
  \hline
  \texttt{indep} ps   & 33.4 & 26.1 & 90.51 & 28.5 & 39.6 & 91.98 & 22.7 & 14.3 & 83.58 & 26.2 (+0.7) & 88.69 (+1.16) \\
  \texttt{dual} ps    & 33.2 & 25.9 & 91.01 & 28.4 & 39.7 & 92.07 & 22.5 & 14.3 & 83.92 & 26.2 (+0.7) & 89.00 (+1.47)\\
  \hline
  \texttt{indep} FT  & 37.1 & 28.6 & 91.52 & 30.1 & 42.3 & 92.25 & 26.5 & 17.1 & 84.86 & 30.3 (+4.8) & 89.54 (+2.01) \\
  \texttt{dual} FT   & 37.1 & 28.0 & 91.53 & 30.0 & 42.3 & 92.43 & 26.0 & 17.0 & 85.03 & 30.1 (+4.6) & 89.66 (+2.13) \\
  \hline
  \texttt{indep} FT+ps & 36.5 & 28.2 & 92.17 & 29.9 & 42.0 & 92.68 & 25.9 & 16.3 & 84.76 & 29.8 (+4.3) & 89.87 (+2.34) \\
  \texttt{dual} FT+ps  & 36.5 & 28.4 & 92.26 & 30.1 & 42.1 & 92.55 & 26.1 & 16.5 & 84.84 & 30.0 (+4.5) & 89.88 (+2.35) \\
  \hline
  \end{tabular}
  }
  \caption{BLEU and similarity scores of multi-target models. Similarity scores (SIM) are computed as the cross-lingual similarity between the two target translations. Pseudo (ps) refers to models trained from scratch with synthetic reference data. FT indicates models fine-tuned from the pre-trained multilingual (\texttt{multi}) model. FT+ps refers to models fine-tuned using synthetic reference data.\label{tab:multi-target}}
\end{table*}

\subsection{Results\label{ssec:mtmt-results}}

We evaluate the performance of models trained only with trilingual data, as well as models pre-trained in a multilingual way. Table~\ref{tab:multi-target} shows that the \texttt{indep} model outperforms the \texttt{base} model in all directions, demonstrating the benefits of jointly training two independent decoders. The same gain is not observed for the \texttt{dual} model, for which results in some directions are even worse than the baseline. One explanation is that dual decoding suffers from a double exposure bias, as errors in both decoders jointly contribute to derail the decoding process. We get back to this issue in Section~\ref{sec:bi-directional}.

To test this, we use the \texttt{base} model to translate source texts $\src$ into targets $\hat\trg^1$ and $\hat\trg^2$, which are then merged with the original data to build a pseudo-trilingual training set. For fair comparison, we only use half of the translations from each target language, yielding a pseudo-trilingual dataset $\{(\src_{1/2}, \hat\trg_{1/2}^1, \trg_{1/2}^2), (\src_{2/2}, \trg_{2/2}^1, \hat\trg_{2/2}^2)\}$ that is as large as the original data.
We see in Table~\ref{tab:multi-target} that these artificial translations almost close the gap between the independent and dual decoders.

Initializing with pre-trained models (Section~\ref{ssec:train}) brings an additional improvement for both methods, thus validating our pre-training procedure (see bottom of Table~\ref{tab:multi-target}). They confirm that dual decoders can be effectively trained, even in the absence of large multi-parallel data. These results also highlight the large gains of fine-tuning on a tri-parallel corpus, which improves our baseline multilingual models by nearly 5~BLEU points on average.

We additionally experiment fine-tuning pre-trained models with the synthetic pseudo-trilingual data. This setting (FT+ps in Table~\ref{tab:multi-target}) does not bring any gain in translation quality: for the  \texttt{indep} model we see a small loss due to training with noisy references; for \texttt{dual}, it seems that mitigating exposure bias is less impactful when starting from well-trained models.

\subsection{Complements and Analysis\label{ssec:mtmt-analysis}}

The value of dual decoding is to ensure that translations $\hat\trg_1$ and $\hat\trg_2$ are more consistent than with independent decoding. To evaluate this, we compute the similarity scores (SIM) between these two translations using LASER.\footnote{\url{https://github.com/facebookresearch/LASER}} As shown in Table~\ref{tab:multi-target}, \texttt{dual} model generate translations that are slightly more similar on average than the \texttt{indep} model: as both translate the same source into the same languages, similarity scores are always quite high.

\begin{table}[!ht]
  \center
  \scalebox{0.85}{
  \begin{tabular}{l|cccccc}
  \hline
  Model  & En-De & En-Fr & \\
  \hline
  \texttt{dual}      & 28.5 & 38.8 \\
  \hline
  \texttt{dual} De+Fr auto  & 28.7 & 39.3 \\
  \texttt{dual} De+Fr ref  & 28.7 & 39.6 \\
  \hline
  \hline
  \texttt{dual} FT   & 30.0 & 42.3 \\
  \hline
  \texttt{dual} FT De wait-3 & 30.2 & 42.4 \\
  \texttt{dual} FT Fr wait-3 & 30.4 & 42.6 \\
  \hline
  \texttt{dual} FT De+Fr auto & 30.0  & 42.5 \\
  \texttt{dual} FT De+Fr ref    & 30.0  & 42.4 \\

  \hline
  \end{tabular}
  }
  \caption{BLEU scores for asynchronous decoding: sequential decoding on the \texttt{dual} model trained from scratch (top), wait-k models fine-tuned on the pre-trained model and sequential decoding on the \texttt{dual} FT model (bottom) for the direction En$\to$De/Fr. Results using sequential decoding for one decoder are obtained in a second decoding pass using either automatic (auto) or reference (ref) translations.\label{tab:wait-k}}
\end{table}

\begin{table*}[!ht]
  \center
  \scalebox{0.85}{
  \begin{tabular}{l|cc|cc|cc|cc|l|l}
  \hline
  Model  & En-De & Cons & En-Fr & Cons & En-Zh & Cons & En-Ja & Cons & Avg BLEU & Avg Cons\\
  \hline
  \texttt{base}       & 28.1 & - & 38.8 & - & 22.2 & - & 13.6 & - & 25.7 & -  \\
  \hline
  \texttt{indep}   & 29.1 & 54.7 & 39.4 & 65.8 & 22.5 & 51.3 & 14.8 & 37.6 & 26.5 (+0.8) & 52.4 \\
  \texttt{dual}    & 25.9 & 88.9 & 36.6 & 90.5 & 20.6 & 86.0 & 4.2  & 68.4 & 21.8 (-3.9) & 83.5 (+31.1)\\
  \hline
  \texttt{indep} pseudo   & 29.0 & 65.7 & 39.9 & 73.3 & 22.9 & 62.0 & 15.6 & 48.6 & 26.9 (+1.2) & 62.4 \\
  \texttt{dual} pseudo    & 28.7 & 83.7 & 38.9 & 89.6 & 23.1 & 80.2 & 15.1 & 67.5 & 26.5 (+0.8) & 80.3 (+17.9) \\
  \hline
  \texttt{indep} pseudo-dup   & 29.3 & 70.9 & 40.5 & 76.3 & 23.5 & 67.6 & 15.8 & 53.7 & 27.3 (+1.6) & 67.1 \\
  \texttt{dual} pseudo-dup    & 29.6 & 83.5 & 40.1 & 89.6 & 23.4 & 78.2 & 15.3 & 70.7 & 27.1 (+1.4) & 80.5 (+13.4) \\
  \hline
  \end{tabular}
  }
  \caption{Results of bi-directional MT models trained with actual data (top) and synthetic data (bottom). The consistency score (Cons) is an averaged BLEU score between the forward and backward translations.\label{tab:bidirect}}
\end{table*}

As explained in Section~\ref{ssec:beam-search}, the dual decoder model is not limited to strictly synchronous generation and accommodates relaxed variants (as well as alternative dependency patterns) where one decoder can start several steps after the other. We fine-tune ``wait-k'' \texttt{dual} models from the pre-trained model with $k=3$ for En$\to$De/Fr and evaluate the effects on performance. As shown in Table~\ref{tab:wait-k}, the BLEU scores are slightly improved for both targets when either side is delayed by $3$ steps. These results suggest that depending on language pairs, the information flow between decoders can be beneficial from a small amount of asynchronicity.

Our implementation also enables to have one decoder finish before the other begins.
We thus experiment a \emph{sequential decoding strategy} (see Section~\ref{ssec:asynchronous}), in which we first compute the complete translation in one target language (with the \texttt{dual} model), then decode the other one. In this case, the second decoding step has access to both the source and the other target sequence. This decoding strategy does not require any additional training and is applied directly during inference. 

We decode both \texttt{dual} and \texttt{dual} FT models with this strategy. Results in Table~\ref{tab:wait-k}, obtained with both automatic and reference translations in one language, show that this technique is able to improve the \texttt{dual} model on both French and German translations, while only slightly improves the French translation for the \texttt{dual} FT model. Sequential decoding with reference in one language provides the other decoder with the ground truth, which therefore alleviates the exposure bias problem suffered by \texttt{dual} models. However, combining results of FT models in Table~\ref{tab:multi-target} and \ref{tab:wait-k}, we see that fine-tuned models are less sensitive to errors made during decoding. This again shows the benefit that \texttt{dual} models actually obtain from pre-trained models.

\section{Bi-directional MT \label{sec:bi-directional}}

Bi-directional MT \citep{Finch09bidirectional,Zhou19synchronous,Liu20agreement} aims to integrate future information in the decoder by jointly translating in the forward (left to right, L2R) and in the backward (right to left, R2L) directions. Another expectation is that the two decoders, having different views of the source, will deliver complementary translations. Dual decoding readily applies in this setting, with one decoder for each direction, with the added benefit of generating more coherent outputs than independent decoders. We evaluate this added consistency by reusing the experimental setting (data, implementation and hyperparameters) of Section~\ref{sec:multi-target}, and by training $4$ bi-directional systems, from English into German, French, Chinese and Japanese. Similar to \citet{Zhou19synchronous}, we output the translation with the highest probability, inverting the translation if the R2L output is picked. 

We first train models on tri-parallel corpora obtained by adding an inverted version of the target sentence to each training sample. In this setting, the \texttt{dual} model again suffers a clear drop of BLEU scores as compared to \texttt{indep} model (Table~\ref{tab:bidirect}). We again attribute this loss to the impact of the exposure bias, as can be seen in Table~\ref{tab:bidirect}, where the loss in BLEU score of the \texttt{dual} system is accompanied by a very large increase in consistency of the outputs (+31.1). We therefore again introduce pseudo-parallel targets, where one of the two targets is automatically generated with the \texttt{base} model. This was also proposed in \cite{Zhou19synchronous,Wang19synchronously,Zhang20synchronous,He21synchronous}.
Similar to the pseudo-data described in Section~\ref{ssec:mtmt-results}, we generate a \texttt{pseudo} dataset
in which each original source sentence occurs just once. This means that the forward and backward training target sentences are not always deterministically related, which forces each decoder to put less trust on tokens from the other direction. We also consider the \texttt{pseudo-dup} data,
in which each source sentence is duplicated, occurring once with the reference data in each direction. Results in Table~\ref{tab:bidirect} show that this method again closes the gap between \texttt{indep} and \texttt{dual}, and yields systems that surpass the baseline by about 1 BLEU point in the \texttt{pseudo} setting, and by 1.5 BLEU point in the \texttt{pseudo-dup} setting.

By computing the BLEU score between the two output translations, we can also evaluate the increment of consistency incurred in dual decoding. These scores are reported in Table~\ref{tab:bidirect} (column \textit{Cons}) and show to a +13.4 BLEU increment when averaged over language pairs, thereby demonstrating the positive impact of dual decoding. 

\section{MT for Code-switched Inputs \label{sec:csw}}

In this section, we turn to a novel task, consisting in translating a code-switched (CSW) sentence (containing fragments from two languages) simultaneously into its two components. An example in Table~\ref{tab:csw-example} for French-English borrowed from \cite{Carpuat14mixed} illustrates this task.

\begin{table}[!ht] 
  \center
  {\footnotesize
    \begin{tabular}{ll} \hline
      \src: & autrement dit, they are getting out of the closet \\
      \trg$^1$: & In other words, they are getting out of the closet \\
      \trg$^2$: & autrement dit, ils sortent du placard \\ 
      \hline
    \end{tabular}
  }
\caption{Dual decoding for a CSW sentence.\label{tab:csw-example}}
\end{table}

Code-switching is an important phenomenon in informal communications between bilingual speakers. It generally consists of short inserts of a secondary language which are embedded within larger fragments in the primary language. When simultaneously translating into these two languages, we expect the following ``copy'' constraint to be satisfied: \emph{every word in the source text should appear in at least one of the two outputs}. 

Our main interest in this experiment is to assess how much dual decoding actually enforces this constraint. As tri-parallel corpora for this task are scarce \citep{Menacer19machine},
we mostly follow \citep{Song19code,Xu21traducir} and automatically generate artificial CSW sentences from regular parallel data. Working with the En-Fr pair, we use the WMT14 En-Fr data to generate training data as well as a CSW version of the \texttt{newstest2014} test set. Approximately half of the test sentences are mostly English with inserts in French, and mostly French with inserts in English for the other half. We use the same pre-training procedure as in Section~\ref{ssec:settings} and evaluate with \texttt{csw-newstest2014} data.

\begin{table}[!ht]
  \center
  \scalebox{0.85}{
  \begin{tabular}{l|cccc}
  \hline
  Model       & \multicolumn{2}{c}{CSW-En} & \multicolumn{2}{c}{CSW-Fr} \\[-2pt]
              &   second  &   primary     & second   &   primary \\
  \hline
  \texttt{base}       & \multicolumn{2}{c}{67.8}    & \multicolumn{2}{c}{67.4} \\[-2pt]
                      & \small{35.5} & \small{97.4} & \small{37.5} & \small{95.3} \\
  \texttt{indep} & \multicolumn{2}{c}{67.7}    & \multicolumn{2}{c}{67.3} \\[-2pt]
                      & \small{35.1} & \small{97.4} & \small{37.1} & \small{95.5} \\
  \texttt{dual}    & \multicolumn{2}{c}{67.7} & \multicolumn{2}{c}{67.5} \\[-2pt]
                      & \small{35.1} & \small{97.5} & \small{37.5} & \small{95.4} \\
  \hline
  \end{tabular}
  }
  \caption{BLEU scores of CSW translation models tested on the \texttt{csw-newstest2014} data that we generated. Small numbers are scores computed separately on the two parts of the test set where the target language is primary or secondary (second).\label{tab:csw}}
\end{table}

Table~\ref{tab:csw} reports overall BLEU scores, as well as scores for the `primary and `secondary' part of the test set for each target language. These results show that \texttt{indep} and \texttt{dual} systems, which are both able to translate French mixed with English and English mixed with French, achieve performance that is comparable to the \texttt{base} model, which, in this experiment, \emph{is made of two distinct Transformer models}, one for each direction.

We also measure how well the constraint expressed above is satisfied. It stipulates that every token in a CSW sentence should be either copied in one language (and translated into the other), or copied in both, which mostly happens for punctuations, numbers or proper names. Our analysis in Table~\ref{tab:csw-copy} shows that the \texttt{base} model is more likely to reproduce the patterns observed in the reference, notably is less likely to generate two copies for a token than the other systems. However, \texttt{indep} and, to a larger extend, \texttt{dual}, are able to reduce the rate of \emph{lost tokens}, i.e.\ of source tokens that are not found in any output. This again shows that the interaction between the two decoders helps to increase the consistency between the two outputs.

\begin{table}[!ht]
  \center
  \scalebox{0.85}{
  \begin{tabular}{l|cccc}
  \hline
  Model       & Exclusive & Both & Punctuations & Lost \\
  \hline
  reference &  81.56    & 8.10 & 10.34 & 0 \\
  \hline
  \texttt{base} & 79.14    & 8.85 & 11.29 & 0.72 \\
  \texttt{indep} & 78.86 & 9.13 & 11.35 & 0.67 \\
  \texttt{dual} & 78.90 & 9.17 & 11.32 & 0.61 \\
  \hline
  \end{tabular}
  }
  \caption{Analysis of the ``copy'' constraint. ``Exclusive'' refers to the percentage of test tokens appearing in only one hypothesis. ``Both'' and ``Punctuations'' are for tokens and punctuations$+$digits appearing in both hypotheses, and ``Lost'' is for tokens not found in either.\label{tab:csw-copy}}
\end{table}

\section{Generating Translation Variants \label{sec:variants}}

As a last application of dual decoding, we study the generation of pairs of consistent translation alternatives, using variation in ``politeness'' as our test bed. We borrow the experimental setting and data of \citet{Sennrich16controlling}.\footnote{\url{http://data.statmt.org/rsennrich/politeness/}} The training set contains $5.58$M sentences pairs, out of which $0.48$M are annotated as polite and $1.06$M as impolite. The rest is deemed neutral.\footnote{See details in Appendix \ref{sec:detail-polite}.}
Using this data, we generate tri-parallel data as follows. We first train a tag-based NMT with politeness control as in \citet{Sennrich16controlling} and use it to predict the polite counterpart of each impolite sentence, and vice-versa. We also include an equivalent number of randomly chosen neutral sentences: for these, the polite and impolite versions are identical. The resulting $3$-way corpus contains $3.07$M sentences.
Similar to the multi-target task (Section~\ref{sec:multi-target}), we fine-tune a pre-trained model with this data until convergence. We use the \texttt{test} data of \citet{Sennrich16controlling} as development set and test our model on the \texttt{testyou} set, which contains 2k sentences with a second-person pronoun \textit{you(r(s(elf)))} in the English source. The annotation tool distributed with the data is used to assess the politeness of the output translations.

Table~\ref{tab:polite} (top) reports the performance of the pre-trained model. \textit{ref} refers to the annotation result of the reference German sentences. \textit{none} is translated without adding any tags to the source text, while \textit{pol} and \textit{imp} are translated with all sentences tagged respectively as \textit{polite} and \textit{impolite}. The \textit{oracle} line is obtained by prefixing each source sentence with the correct tag. These results show the effectiveness of side constraints for the generation of variants: for both polite and impolite categories, the pre-trained model generates translations that mostly satisfy the desired requirement. 

\begin{table}[!ht]
  \center
  \scalebox{0.85}{
  \begin{tabular}{l|l|ccc|c}
  \hline
  Model & Tag         & neutral & pol & imp & BLEU \\
  \hline
        & ref &   438   &   525  &   1037   & - \\
  \hline
  \texttt{pre-} & none        &   1914  &   16   &   70     & 17.7 \\
  \texttt{train}& pol      &  479    &  1518  &   3      & 20.9 \\
                     & imp    &  22     &    0   &   1978   & 24.1 \\
                     & oracle      &  551    &  406   &   1043   & 30.2 \\
  \hline
  \hline
        & Dec     & neutral & pol & imp & BLEU \\
  \hline
  \texttt{indep} & pol      &  528    &  1470  &   2      & 21.0 \\
                      & imp    &  82     &    0   &   1918   & 24.4 \\
  \hline
  \texttt{dual}    & pol      &  541    &  1457  &   2      & 21.3 \\
                      & imp    &  97     &    0   &   1903   & 24.3 \\
  \hline
  seq imp     & pol      &  531    &  1467  &   2      & 21.2 \\
  seq pol     & imp     &  84     &  0     &   1916   & 24.4 \\
  \hline
  \end{tabular}
  }
  \caption{Results of politeness MT models. Tags are used for the \texttt{pre-train} model to generate the desired variant. Decoders (Dec) of \texttt{indep} and \texttt{dual} compute two translations in one decoding step, while the results using sequential decoding for one decoder are obtained with the $2$-step procedure of Section~\ref{ssec:asynchronous}.\label{tab:polite}}
\end{table}

Results of the fine-tuned dual decoder models are in Table~\ref{tab:polite} (bottom): we see that both models are very close and generate more neutral translations and also slightly improve the BLEU scores compared to the pre-trained model.

As discussed in Section~\ref{ssec:asynchronous}, our dual decoder model can delay one decoder until the other is finished. We redo the same sequential decoding procedure as in Section~\ref{ssec:mtmt-analysis}.
Results in Table~\ref{tab:polite} (bottom) indicate that given the full translation of impolite variations, the \texttt{dual} model tends to generate less neutral sentences but more polite ones. The same phenomenon is also observed in the other direction. This implies that the output variations can be better controlled with sequential decoding.

\section{Related Work \label{sec:related}}
The variety of applications considered here makes it difficult to give a thorough analysis of all the related work, and we only mention the most significant landmarks.

\paragraph{Multi-source / Multi-target Machine Translation} Multi-source MT was studied in the framework of SMT, considering with a tight integration (in the decoder), or a late integration (by combining multiple hypotheses obtained with different sources). This idea was revisited in the Neural framework \citep{Zoph16multi-source,Liu20meet}. Setting multilingual MT aside \cite{Dabre20survey}, studies of the multi-target case are comparatively rarer \citep{Neubig15multi}. Notable references are \cite{Dong15multitask}, which introduces a multi-task framework; \citep{Wang18three}, which studies ways to strengthen a basic multilingual decoder; while closer to our work, \citet{Wang19synchronously} consider a dual decoder relying on dual self-attention mechanism. Related techniques have also been used to simultaneously generate a transcript and a translation for a spoken input \citep{Anastasopoulos18tied,Le20dual} and to generate consistent caption and subtitle for an audio source \citep{Karakanta21flexibility}.

\paragraph{Bi-directional Decoding} is an old idea from the statistical MT era \citep{Watanabe02bidirectional,Finch09bidirectional}. Instantiations of these techniques for NMT are in \citep{Zhang18asynchronous,Su19exploiting}, where asynchronous search techniques are considered; and in \cite{Zhou19synchronous,Wang19synchronously,Zhang20synchronous} where, similar to our work, various ways to enforce a tighter interaction between directions are considered in synchronous search, while \citet{Liu20agreement} also study ways to increase the agreement between L2R and R2L directions. More recently, \citep{He21synchronous} combines multi-target and bi-directional decoding within a single architecture, where, in each layer and block, all cross-attentions are combined with a single hidden state; four softmax layers are used for the output symbols in a proposal that creates an even stronger dependency between decoders than what we consider here.

\paragraph{Code-switching} is an important linguistic phenomenon in bilingual communities that is getting momentum within the natural language processing communities \citep{Sitaram19survey}. Several tasks have been considered:  token-level language identification \citep{Samih16multilingual}, Language Modeling \citep{Winata19code}, Named Entity Recognition \citep{Aguilar18overview}, Part-of-Speech tagging \citep{Ball18partofspeech} and Sentiment Analysis \citep{Patwa20semeval}. Machine Translation for CSW texts is considered in \citep{Menacer19machine}.

\section{Conclusion and Future Work}
In this paper, we have explored various possible implementations of dual decoding, as a way to generate pairs of consistent translation. Dual decoding can be viewed as a tight form of multi-task learning, and, as we have seen, can be effectively trained using actual or partly artificial data; it can also directly benefit from pre-trained models. Considering four applications of MT, we have observed that dual decoding was prone to exposure bias in the two decoders, and we have proposed practical remedies. Using these, we have achieved BLEU scores that match those of a simple multi-task learners, and display an increased level of consistency.

In our future work, we plan to consider other strategies, such as scheduled sampling \citep{Bengio15scheduled,Mihaylova19scheduled}, to mitigate the exposure bias. Another area where we seek to improve is the relaxation of strict synchronicity in decoding. We finally wish to study more applications of this technique, notably to generate controlled variation: controlling gender variation \citep{Zmigrod19counterfactual} or more complex form of formality levels, as in \citep{Niu20controlling}, are obvious candidates.

\section*{Acknowledgements \label{sec:acknowledgements}}
This work was granted access to the HPC resources of IDRIS under the allocation 2021-[AD011011580R1] made by GENCI. The authors wish to thank Josep Crego for his comments and discussions. We would also like to thank the anonymous reviewers for their valuable suggestions. The first author is partly funded by Systran and by a grant from Région Ile-de-France. 

\bibliography{multi_decoding}

\begin{thebibliography}{52}
\expandafter\ifx\csname natexlab\endcsname\relax\def\natexlab#1{#1}\fi

\bibitem[{Aguilar et~al.(2018)Aguilar, AlGhamdi, Soto, Diab, Hirschberg, and
  Solorio}]{Aguilar18overview}
Gustavo Aguilar, Fahad AlGhamdi, Victor Soto, Mona Diab, Julia Hirschberg, and
  Thamar Solorio. 2018.
\newblock \href {https://doi.org/10.18653/v1/W18-3219} {Named entity
  recognition on code-switched data: Overview of the {CALCS} 2018 shared task}.
\newblock In \emph{Proceedings of the Third Workshop on Computational
  Approaches to Linguistic Code-Switching}, pages 138--147, Melbourne,
  Australia. Association for Computational Linguistics.

\bibitem[{Anastasopoulos and Chiang(2018)}]{Anastasopoulos18tied}
Antonios Anastasopoulos and David Chiang. 2018.
\newblock \href {https://doi.org/10.18653/v1/N18-1008} {Tied multitask learning
  for neural speech translation}.
\newblock In \emph{Proceedings of the 2018 Conference of the North {A}merican
  Chapter of the Association for Computational Linguistics: Human Language
  Technologies, Volume 1 (Long Papers)}, pages 82--91, New Orleans, Louisiana.
  Association for Computational Linguistics.

\bibitem[{Bahdanau et~al.(2015)Bahdanau, Cho, and Bengio}]{Bahdanau15neural}
Dzmitry Bahdanau, Kyunghyun Cho, and Yoshua Bengio. 2015.
\newblock \href {http://arxiv.org/abs/1409.0473} {Neural machine translation by
  jointly learning to align and translate}.
\newblock In \emph{3rd International Conference on Learning Representations,
  {ICLR} 2015, San Diego, CA, USA, May 7-9, 2015, Conference Track
  Proceedings}.

\bibitem[{Ball and Garrette(2018)}]{Ball18partofspeech}
Kelsey Ball and Dan Garrette. 2018.
\newblock \href {https://doi.org/10.18653/v1/D18-1347} {Part-of-speech tagging
  for code-switched, transliterated texts without explicit language
  identification}.
\newblock In \emph{Proceedings of the 2018 Conference on Empirical Methods in
  Natural Language Processing}, pages 3084--3089, Brussels, Belgium.
  Association for Computational Linguistics.

\bibitem[{Bengio et~al.(2015)Bengio, Vinyals, Jaitly, and
  Shazeer}]{Bengio15scheduled}
Samy Bengio, Oriol Vinyals, Navdeep Jaitly, and Noam Shazeer. 2015.
\newblock \href
  {https://proceedings.neurips.cc/paper/2015/file/e995f98d56967d946471af29d7bf99f1-Paper.pdf}
  {Scheduled sampling for sequence prediction with recurrent neural networks}.
\newblock In \emph{Advances in Neural Information Processing Systems},
  volume~28. Curran Associates, Inc.

\bibitem[{Bojanowski et~al.(2017)Bojanowski, Grave, Joulin, and
  Mikolov}]{Bojanowski17enriching}
Piotr Bojanowski, Edouard Grave, Armand Joulin, and Tomas Mikolov. 2017.
\newblock \href {https://doi.org/10.1162/tacl_a_00051} {Enriching word vectors
  with subword information}.
\newblock \emph{Transactions of the Association for Computational Linguistics},
  5:135--146.

\bibitem[{Carpuat(2014)}]{Carpuat14mixed}
Marine Carpuat. 2014.
\newblock \href {https://doi.org/10.3115/v1/W14-3913} {Mixed language and
  code-switching in the {C}anadian hansard}.
\newblock In \emph{Proceedings of the First Workshop on Computational
  Approaches to Code Switching}, pages 107--115, Doha, Qatar. Association for
  Computational Linguistics.

\bibitem[{Cettolo et~al.(2012)Cettolo, Girardi, and Federico}]{Cettolo12wit3}
Mauro Cettolo, Christian Girardi, and Marcello Federico. 2012.
\newblock \href {https://www.aclweb.org/anthology/2012.eamt-1.60} {{WIT}3: Web
  inventory of transcribed and translated talks}.
\newblock In \emph{Proceedings of the 16th Annual conference of the European
  Association for Machine Translation}, pages 261--268, Trento, Italy. European
  Association for Machine Translation.

\bibitem[{Cho et~al.(2014)Cho, van Merri{\"e}nboer, Bahdanau, and
  Bengio}]{Cho14properties}
Kyunghyun Cho, Bart van Merri{\"e}nboer, Dzmitry Bahdanau, and Yoshua Bengio.
  2014.
\newblock \href {https://doi.org/10.3115/v1/W14-4012} {On the properties of
  neural machine translation: Encoder{--}decoder approaches}.
\newblock In \emph{Proceedings of {SSST}-8, Eighth Workshop on Syntax,
  Semantics and Structure in Statistical Translation}, pages 103--111, Doha,
  Qatar. Association for Computational Linguistics.

\bibitem[{Crego et~al.(2005)Crego, Mariño, and Gispert}]{Crego05reordered}
Josep~M. Crego, José~B. Mariño, and Adrià~De Gispert. 2005.
\newblock {Reordered search, and tuple unfolding for Ngram-based SMT}.
\newblock In \emph{In Proceedings of the MT Summit X}, pages 283--289.

\bibitem[{Dabre et~al.(2020)Dabre, Chu, and Kunchukuttan}]{Dabre20survey}
Raj Dabre, Chenhui Chu, and Anoop Kunchukuttan. 2020.
\newblock \href {https://doi.org/10.1145/3406095} {A survey of multilingual
  neural machine translation}.
\newblock \emph{ACM Comput. Surv.}, 53(5).

\bibitem[{Dong et~al.(2015)Dong, Wu, He, Yu, and Wang}]{Dong15multitask}
Daxiang Dong, Hua Wu, Wei He, Dianhai Yu, and Haifeng Wang. 2015.
\newblock \href {https://doi.org/10.3115/v1/P15-1166} {Multi-task learning for
  multiple language translation}.
\newblock In \emph{Proceedings of the 53rd Annual Meeting of the Association
  for Computational Linguistics and the 7th International Joint Conference on
  Natural Language Processing (Volume 1: Long Papers)}, pages 1723--1732,
  Beijing, China. Association for Computational Linguistics.

\bibitem[{Dyer et~al.(2013)Dyer, Chahuneau, and Smith}]{Dyer13simple}
Chris Dyer, Victor Chahuneau, and Noah~A. Smith. 2013.
\newblock \href {https://www.aclweb.org/anthology/N13-1073} {A simple, fast,
  and effective reparameterization of {IBM} model 2}.
\newblock In \emph{Proceedings of the 2013 Conference of the North {A}merican
  Chapter of the Association for Computational Linguistics: Human Language
  Technologies}, pages 644--648, Atlanta, Georgia. Association for
  Computational Linguistics.

\bibitem[{Elbayad et~al.(2020)Elbayad, Besacier, and Verbeek}]{Elbayad20waitk}
Maha Elbayad, Laurent Besacier, and Jakob Verbeek. 2020.
\newblock \href {https://doi.org/10.21437/Interspeech.2020-1241} {{Efficient
  Wait-k Models for Simultaneous Machine Translation}}.
\newblock In \emph{{Interspeech 2020 - Conference of the International Speech
  Communication Association}}, pages 1461--1465, Shangai (Virtual Conf), China.

\bibitem[{Finch and Sumita(2009)}]{Finch09bidirectional}
Andrew Finch and Eiichiro Sumita. 2009.
\newblock \href {https://www.aclweb.org/anthology/D09-1117} {Bidirectional
  phrase-based statistical machine translation}.
\newblock In \emph{Proceedings of the 2009 Conference on Empirical Methods in
  Natural Language Processing}, pages 1124--1132, Singapore. Association for
  Computational Linguistics.

\bibitem[{Firat et~al.(2016)Firat, Cho, and Bengio}]{Firat16multiway}
Orhan Firat, Kyunghyun Cho, and Yoshua Bengio. 2016.
\newblock \href {https://doi.org/10.18653/v1/N16-1101} {Multi-way, multilingual
  neural machine translation with a shared attention mechanism}.
\newblock In \emph{Proceedings of the 2016 Conference of the North American
  Chapter of the Association for Computational Linguistics: Human Language
  Technologies}, pages 866--875. Association for Computational Linguistics.

\bibitem[{Ha et~al.(2016)Ha, Niehues, and Waibel}]{Ha16towards}
Thanh-He Ha, Jan Niehues, and Alex Waibel. 2016.
\newblock Toward multilingual neural machine translation with universal encoder
  and decoder.
\newblock In \emph{Proceedings of the 13th International Workshop on Spoken
  Language Translation}, IWSLT 2016, Vancouver, Canada.

\bibitem[{He et~al.(2021)He, Wang, Yu, Zhao, Zhang, and Zong}]{He21synchronous}
Hao He, Qian Wang, Zhipeng Yu, Yang Zhao, Jiajun Zhang, and Chengqing Zong.
  2021.
\newblock \href {https://ojs.aaai.org/index.php/AAAI/article/view/17535}
  {Synchronous interactive decoding for multilingual neural machine
  translation}.
\newblock In \emph{Proceedings of the AAAI Conference on Artificial
  Intelligence}, volume~35, pages 12981--12988.

\bibitem[{Inan et~al.(2017)Inan, Khosravi, and Socher}]{Inan17tying}
Hakan Inan, Khashayar Khosravi, and Richard Socher. 2017.
\newblock \href {https://openreview.net/forum?id=r1aPbsFle} {Tying word vectors
  and word classifiers: {A} loss framework for language modeling}.
\newblock In \emph{5th International Conference on Learning Representations,
  {ICLR} 2017, Toulon, France, April 24-26, 2017, Conference Track
  Proceedings}. OpenReview.net.

\bibitem[{Johnson et~al.(2017)Johnson, Schuster, Le, Krikun, Wu, Thorat,
  Vi{\'e}gas, Wattenberg, Corrado, Hughes, and Dean}]{Johnson17googles}
Melvin Johnson, Mike Schuster, Quoc~V. Le, Maxim Krikun, Zhifeng Wu,
  Yonghui~andhen, Nikhil Thorat, Fernanda Vi{\'e}gas, Martin Wattenberg, Greg
  Corrado, Macduff Hughes, and Jeffrey Dean. 2017.
\newblock \href {https://doi.org/10.1162/tacl_a_00065} {{G}oogle{'}s
  multilingual neural machine translation system: Enabling zero-shot
  translation}.
\newblock \emph{Transactions of the Association for Computational Linguistics},
  5:339--351.

\bibitem[{Karakanta et~al.(2021)Karakanta, Gaido, Negri, and
  Turchi}]{Karakanta21flexibility}
Alina Karakanta, Marco Gaido, Matteo Negri, and Marco Turchi. 2021.
\newblock \href {https://doi.org/10.18653/v1/2021.iwslt-1.26} {Between
  flexibility and consistency: Joint generation of captions and subtitles}.
\newblock In \emph{Proceedings of the 18th International Conference on Spoken
  Language Translation (IWSLT 2021)}, pages 215--225, Bangkok, Thailand
  (online). Association for Computational Linguistics.

\bibitem[{Le et~al.(2020)Le, Pino, Wang, Gu, Schwab, and Besacier}]{Le20dual}
Hang Le, Juan Pino, Changhan Wang, Jiatao Gu, Didier Schwab, and Laurent
  Besacier. 2020.
\newblock \href {https://doi.org/10.18653/v1/2020.coling-main.314}
  {Dual-decoder transformer for joint automatic speech recognition and
  multilingual speech translation}.
\newblock In \emph{Proceedings of the 28th International Conference on
  Computational Linguistics}, pages 3520--3533, Barcelona, Spain (Online).
  International Committee on Computational Linguistics.

\bibitem[{Liu et~al.(2020{\natexlab{a}})Liu, Luo, Ao, Song, Xu, and
  Ye}]{Liu20meet}
Jianfeng Liu, Ling Luo, Xiang Ao, Yan Song, Haoran Xu, and Jian Ye.
  2020{\natexlab{a}}.
\newblock \href {https://doi.org/10.18653/v1/2020.coling-main.97} {Meet changes
  with constancy: Learning invariance in multi-source translation}.
\newblock In \emph{Proceedings of the 28th International Conference on
  Computational Linguistics}, pages 1122--1132, Barcelona, Spain (Online).
  International Committee on Computational Linguistics.

\bibitem[{Liu et~al.(2020{\natexlab{b}})Liu, Finch, Utiyama, and
  Sumita}]{Liu20agreement}
Lemao Liu, Andrew Finch, Masao Utiyama, and Eiichiro Sumita.
  2020{\natexlab{b}}.
\newblock \href {https://www.jair.org/index.php/jair/article/view/12008}
  {Agreement on target-bidirectional recurrent neural networks for
  sequence-to-sequence learning}.
\newblock \emph{Journal of Artificial Intelligence Research}, 67:581--606.

\bibitem[{Menacer et~al.(2019)Menacer, Langlois, Jouvet, Fohr, Mella, and
  Sma{\"i}li}]{Menacer19machine}
Mohamed Menacer, David Langlois, Denis Jouvet, Dominique Fohr, Odile Mella, and
  Kamel Sma{\"i}li. 2019.
\newblock \href {https://hal.archives-ouvertes.fr/hal-02106010} {{Machine
  Translation on a parallel Code-Switched Corpus}}.
\newblock In \emph{{Canadian AI 2019 - 32nd Conference on Canadian Artificial
  Intelligence}}, Lecture Notes in Artificial Intelligence, Ontario, Canada.

\bibitem[{Mihaylova and Martins(2019)}]{Mihaylova19scheduled}
Tsvetomila Mihaylova and Andr{\'e} F.~T. Martins. 2019.
\newblock \href {https://doi.org/10.18653/v1/P19-2049} {Scheduled sampling for
  transformers}.
\newblock In \emph{Proceedings of the 57th Annual Meeting of the Association
  for Computational Linguistics: Student Research Workshop}, pages 351--356,
  Florence, Italy. Association for Computational Linguistics.

\bibitem[{Neubig et~al.(2015)Neubig, Arthur, and Duh}]{Neubig15multi}
Graham Neubig, Philip Arthur, and Kevin Duh. 2015.
\newblock \href {https://doi.org/10.3115/v1/N15-1033} {Multi-target machine
  translation with multi-synchronous context-free grammars}.
\newblock In \emph{Proceedings of the 2015 Conference of the North {A}merican
  Chapter of the Association for Computational Linguistics: Human Language
  Technologies}, pages 293--302, Denver, Colorado. Association for
  Computational Linguistics.

\bibitem[{Niu and Carpuat(2020)}]{Niu20controlling}
Xing Niu and Marine Carpuat. 2020.
\newblock \href {https://doi.org/10.1609/aaai.v34i05.6379} {Controlling neural
  machine translation formality with synthetic supervision}.
\newblock In \emph{Proceedings of the AAAI Conference on Artificial
  Intelligence}, volume~34, pages 8568--8575.

\bibitem[{Och and Ney(2001)}]{Och01multisource}
Franz~Josef Och and Hermann Ney. 2001.
\newblock Statistical multi-source translation.
\newblock In \emph{Proceedings of MT Summit}, Santiago de Compostela, Spain.

\bibitem[{Ott et~al.(2019)Ott, Edunov, Baevski, Fan, Gross, Ng, Grangier, and
  Auli}]{Ott19fairseq}
Myle Ott, Sergey Edunov, Alexei Baevski, Angela Fan, Sam Gross, Nathan Ng,
  David Grangier, and Michael Auli. 2019.
\newblock \href {https://doi.org/10.18653/v1/N19-4009} {fairseq: A fast,
  extensible toolkit for sequence modeling}.
\newblock In \emph{Proceedings of the 2019 Conference of the North {A}merican
  Chapter of the Association for Computational Linguistics (Demonstrations)},
  pages 48--53, Minneapolis, Minnesota. Association for Computational
  Linguistics.

\bibitem[{Patwa et~al.(2020)Patwa, Aguilar, Kar, Pandey, PYKL, Gamb{\"a}ck,
  Chakraborty, Solorio, and Das}]{Patwa20semeval}
Parth Patwa, Gustavo Aguilar, Sudipta Kar, Suraj Pandey, Srinivas PYKL,
  Bj{\"o}rn Gamb{\"a}ck, Tanmoy Chakraborty, Thamar Solorio, and Amitava Das.
  2020.
\newblock \href {https://www.aclweb.org/anthology/2020.semeval-1.100}
  {{S}em{E}val-2020 task 9: Overview of sentiment analysis of code-mixed
  tweets}.
\newblock In \emph{Proceedings of the Fourteenth Workshop on Semantic
  Evaluation}, pages 774--790, Barcelona (online). International Committee for
  Computational Linguistics.

\bibitem[{Post(2018)}]{Post18sacrebleu}
Matt Post. 2018.
\newblock \href {https://doi.org/10.18653/v1/W18-6319} {A call for clarity in
  reporting {BLEU} scores}.
\newblock In \emph{Proceedings of the Third Conference on Machine Translation:
  Research Papers}, pages 186--191, Brussels, Belgium. Association for
  Computational Linguistics.

\bibitem[{Press and Wolf(2017)}]{Press17using}
Ofir Press and Lior Wolf. 2017.
\newblock \href {https://aclanthology.org/E17-2025} {Using the output embedding
  to improve language models}.
\newblock In \emph{Proceedings of the 15th Conference of the {E}uropean Chapter
  of the Association for Computational Linguistics: Volume 2, Short Papers},
  pages 157--163, Valencia, Spain. Association for Computational Linguistics.

\bibitem[{Samih et~al.(2016)Samih, Maharjan, Attia, Kallmeyer, and
  Solorio}]{Samih16multilingual}
Younes Samih, Suraj Maharjan, Mohammed Attia, Laura Kallmeyer, and Thamar
  Solorio. 2016.
\newblock \href {https://doi.org/10.18653/v1/W16-5806} {Multilingual
  code-switching identification via {LSTM} recurrent neural networks}.
\newblock In \emph{Proceedings of the Second Workshop on Computational
  Approaches to Code Switching}, pages 50--59, Austin, Texas. Association for
  Computational Linguistics.

\bibitem[{Schwartz(2008)}]{Schwartz08multisource}
Lane Schwartz. 2008.
\newblock Multi-source translation methods.
\newblock In \emph{{MT at work}: Proceedings of the Eighth Conference of the
  Association for Machine Translation in the Americas}, pages 279--288,
  Waikiki, Hawaii.

\bibitem[{Sennrich et~al.(2016{\natexlab{a}})Sennrich, Haddow, and
  Birch}]{Sennrich16controlling}
Rico Sennrich, Barry Haddow, and Alexandra Birch. 2016{\natexlab{a}}.
\newblock \href {https://doi.org/10.18653/v1/N16-1005} {Controlling politeness
  in neural machine translation via side constraints}.
\newblock In \emph{Proceedings of the 2016 Conference of the North {A}merican
  Chapter of the Association for Computational Linguistics: Human Language
  Technologies}, pages 35--40, San Diego, California. Association for
  Computational Linguistics.

\bibitem[{Sennrich et~al.(2016{\natexlab{b}})Sennrich, Haddow, and
  Birch}]{Sennrich16BPE}
Rico Sennrich, Barry Haddow, and Alexandra Birch. 2016{\natexlab{b}}.
\newblock \href {https://doi.org/10.18653/v1/P16-1162} {Neural machine
  translation of rare words with subword units}.
\newblock In \emph{Proceedings of the 54th Annual Meeting of the Association
  for Computational Linguistics (Volume 1: Long Papers)}, pages 1715--1725,
  Berlin, Germany. Association for Computational Linguistics.

\bibitem[{Sitaram et~al.(2019)Sitaram, Chandu, Rallabandi, and
  Black}]{Sitaram19survey}
Sunayana Sitaram, Khyathi~Raghavi Chandu, Sai~Krishna Rallabandi, and Alan~W.
  Black. 2019.
\newblock \href {http://arxiv.org/abs/1904.00784} {A survey of code-switched
  speech and language processing}.
\newblock \emph{CoRR}, abs/1904.00784.

\bibitem[{Song et~al.(2019)Song, Zhang, Yu, Luo, Wang, and Zhang}]{Song19code}
Kai Song, Yue Zhang, Heng Yu, Weihua Luo, Kun Wang, and Min Zhang. 2019.
\newblock \href {https://doi.org/10.18653/v1/N19-1044} {Code-switching for
  enhancing {NMT} with pre-specified translation}.
\newblock In \emph{Proceedings of the 2019 Conference of the North {A}merican
  Chapter of the Association for Computational Linguistics: Human Language
  Technologies, Volume 1 (Long and Short Papers)}, pages 449--459, Minneapolis,
  Minnesota. Association for Computational Linguistics.

\bibitem[{Su et~al.(2019)Su, Zhang, Lin, Qin, Yao, and Liu}]{Su19exploiting}
Jinsong Su, Xiangwen Zhang, Qian Lin, Yue Qin, Junfeng Yao, and Yang Liu. 2019.
\newblock \href {https://doi.org/https://doi.org/10.1016/j.artint.2019.103168}
  {Exploiting reverse target-side contexts for neural machine translation via
  asynchronous bidirectional decoding}.
\newblock \emph{Artificial Intelligence}, 277:103168.

\bibitem[{Sutskever et~al.(2014)Sutskever, Vinyals, and
  Le}]{Sutskever14sequence}
Ilya Sutskever, Oriol Vinyals, and Quoc~V Le. 2014.
\newblock \href
  {https://proceedings.neurips.cc/paper/2014/file/a14ac55a4f27472c5d894ec1c3c743d2-Paper.pdf}
  {Sequence to sequence learning with neural networks}.
\newblock In \emph{Advances in Neural Information Processing Systems},
  volume~27, pages 3104--3112. Curran Associates, Inc.

\bibitem[{Vaswani et~al.(2017)Vaswani, Shazeer, Parmar, Uszkoreit, Jones,
  Gomez, Kaiser, and Polosukhin}]{Vaswani17attention}
Ashish Vaswani, Noam Shazeer, Niki Parmar, Jakob Uszkoreit, Llion Jones,
  Aidan~N Gomez, \L~ukasz Kaiser, and Illia Polosukhin. 2017.
\newblock \href
  {http://papers.nips.cc/paper/7181-attention-is-all-you-need.pdf} {Attention
  is all you need}.
\newblock In I.~Guyon, U.~V. Luxburg, S.~Bengio, H.~Wallach, R.~Fergus,
  S.~Vishwanathan, and R.~Garnett, editors, \emph{Advances in Neural
  Information Processing Systems 30}, pages 5998--6008. Curran Associates, Inc.

\bibitem[{Wang et~al.(2018)Wang, Zhang, Zhai, Xu, and Zong}]{Wang18three}
Yining Wang, Jiajun Zhang, Feifei Zhai, Jingfang Xu, and Chengqing Zong. 2018.
\newblock \href {https://doi.org/10.18653/v1/D18-1326} {Three strategies to
  improve one-to-many multilingual translation}.
\newblock In \emph{Proceedings of the 2018 Conference on Empirical Methods in
  Natural Language Processing}, pages 2955--2960, Brussels, Belgium.
  Association for Computational Linguistics.

\bibitem[{Wang et~al.(2019)Wang, Zhang, Zhou, Liu, and
  Zong}]{Wang19synchronously}
Yining Wang, Jiajun Zhang, Long Zhou, Yuchen Liu, and Chengqing Zong. 2019.
\newblock \href {https://doi.org/10.18653/v1/D19-1330} {Synchronously
  generating two languages with interactive decoding}.
\newblock In \emph{Proceedings of the 2019 Conference on Empirical Methods in
  Natural Language Processing and the 9th International Joint Conference on
  Natural Language Processing (EMNLP-IJCNLP)}, pages 3350--3355, Hong Kong,
  China. Association for Computational Linguistics.

\bibitem[{Watanabe and Sumita(2002)}]{Watanabe02bidirectional}
Taro Watanabe and Eiichiro Sumita. 2002.
\newblock \href {https://www.aclweb.org/anthology/C02-1050} {Bidirectional
  decoding for statistical machine translation}.
\newblock In \emph{{COLING} 2002: The 19th International Conference on
  Computational Linguistics}.

\bibitem[{Winata et~al.(2019)Winata, Madotto, Wu, and Fung}]{Winata19code}
Genta~Indra Winata, Andrea Madotto, Chien-Sheng Wu, and Pascale Fung. 2019.
\newblock \href {https://doi.org/10.18653/v1/K19-1026} {Code-switched language
  models using neural based synthetic data from parallel sentences}.
\newblock In \emph{Proceedings of the 23rd Conference on Computational Natural
  Language Learning (CoNLL)}, pages 271--280, Hong Kong, China. Association for
  Computational Linguistics.

\bibitem[{Xu and Yvon(2021)}]{Xu21traducir}
Jitao Xu and Fran{\c{c}}ois Yvon. 2021.
\newblock \href {https://doi.org/10.18653/v1/2021.calcs-1.11} {Can you traducir
  this? {M}achine translation for code-switched input}.
\newblock In \emph{Proceedings of the Fifth Workshop on Computational
  Approaches to Linguistic Code-Switching}, pages 84--94, Online. Association
  for Computational Linguistics.

\bibitem[{Zhang et~al.(2020)Zhang, Zhou, Zhao, and Zong}]{Zhang20synchronous}
Jiajun Zhang, Long Zhou, Yang Zhao, and Chengqing Zong. 2020.
\newblock \href {https://doi.org/https://doi.org/10.1016/j.artint.2020.103234}
  {Synchronous bidirectional inference for neural sequence generation}.
\newblock \emph{Artificial Intelligence}, 281:103234.

\bibitem[{Zhang et~al.(2018)Zhang, Su, Qin, Liu, Ji, and
  Wang}]{Zhang18asynchronous}
Xiangwen Zhang, Jinsong Su, Yue Qin, Yang Liu, Rongrong Ji, and Hongji Wang.
  2018.
\newblock \href
  {https://www.aaai.org/ocs/index.php/AAAI/AAAI18/paper/view/16784}
  {Asynchronous bidirectional decoding for neural machine translation}.
\newblock In \emph{Proceedings of the Thirty-Second {AAAI} Conference on
  Artificial Intelligence, (AAAI-18), the 30th innovative Applications of
  Artificial Intelligence (IAAI-18), and the 8th {AAAI} Symposium on
  Educational Advances in Artificial Intelligence (EAAI-18), New Orleans,
  Louisiana, USA, February 2-7, 2018}, pages 5698--5705. {AAAI} Press.

\bibitem[{Zhou et~al.(2019)Zhou, Zhang, and Zong}]{Zhou19synchronous}
Long Zhou, Jiajun Zhang, and Chengqing Zong. 2019.
\newblock \href {https://doi.org/10.1162/tacl_a_00256} {Synchronous
  bidirectional neural machine translation}.
\newblock \emph{Transactions of the Association for Computational Linguistics},
  7:91--105.

\bibitem[{Zmigrod et~al.(2019)Zmigrod, Mielke, Wallach, and
  Cotterell}]{Zmigrod19counterfactual}
Ran Zmigrod, Sabrina~J. Mielke, Hanna Wallach, and Ryan Cotterell. 2019.
\newblock \href {https://doi.org/10.18653/v1/P19-1161} {Counterfactual data
  augmentation for mitigating gender stereotypes in languages with rich
  morphology}.
\newblock In \emph{Proceedings of the 57th Annual Meeting of the Association
  for Computational Linguistics}, pages 1651--1661, Florence, Italy.
  Association for Computational Linguistics.

\bibitem[{Zoph and Knight(2016)}]{Zoph16multi-source}
Barret Zoph and Kevin Knight. 2016.
\newblock \href {http://www.aclweb.org/anthology/N16-1004} {Multi-source neural
  translation}.
\newblock In \emph{Proceedings of the 2016 Conference of the North American
  Chapter of the Association for Computational Linguistics: Human Language
  Technologies}, pages 30--34, San Diego, California. Association for
  Computational Linguistics.

\end{thebibliography}
\bibliographystyle{acl_natbib}

\clearpage

\appendix

\section{Details of Data for Multi-target and Bi-directional Machine Translation\label{sec:detail-mtmt}}

We use the IWSLT17 dataset as training data. We use IWSLT17.TED.tst2012 and IWSLT17.TED.tst2013 as development set and test our model on IWSLT17.TED.tst2014 and IWSLT17.TED.tst2015. The original data is not entirely multi-parallel. Therefore, we extract the shared English sentences from En-De and En-Fr data with the corresponding translation to build a truly trilingual corpus. The En$\to$Zh/Ja trilingual data is built similarly. Table~\ref{tab:data-full} summarizes the statistics for the trilingual training and test data.

\begin{table}[!ht]
  \center
  \scalebox{0.9}{
  \begin{tabular}{l|cc|c}
  \hline
       & Original De & Original Fr & 3-way \\
  \hline
  Train &    209522  &   236653    &  205397   \\
  Dev   &    2693    &    3083     &  2468     \\
  tst2014 &    1305  &    1306     &  1168     \\
  tst2015 &    1080  &    1210     &  1074     \\
  \hline
       & Original Zh & Original Ja & 3-way \\
  \hline
  Train    &     235078  &    226834   & 213090    \\
  Dev      &     3064    &     3024    & 2837      \\
  tst2014  &     1297    &     1285    & 1214      \\
  tst2015  &     1205    &     1194    & 1132      \\
  \hline
  \end{tabular}
  }
  \caption{Statistics of extracted trilingual IWSLT data. English is used to extract trilingual sentences therefore not shown in this table.\label{tab:data-full}}
\end{table}

We use WMT20 De-En, De-Fr, En-Zh, En-Ja and WMT14 En-Fr bilingual data for our pre-training experiments. For De-En, De-Fr and En-Fr, we discard the ParaCrawl data and use all the rest. For En-Zh, we only use News Commentary, Wiki Titles, CCMT corpus and WikiMatrix data. For En-Ja, we use all data except ParaCrawl and TED talks. The latter is our trilingual data that we do not use in our pre-training stage. For all WMT data, we discard sentence pairs with invalid language tag as computed by \texttt{fasttext} language identification model \footnote{\url{https://dl.fbaipublicfiles.com/fasttext/supervised-models/lid.176.bin}} \citep{Bojanowski17enriching}. Detailed statistics for the WMT data that we have actually used for each language pair are in Table~\ref{tab:pre-train}.

\begin{table}[!ht]
  \center
  \scalebox{0.9}{
  \begin{tabular}{l|c}
  \hline
  Language pair & \#Sentence(M) \\
  \hline
  En-De & 11.52 \\
  En-Fr & 33.90 \\
  De-Fr & 5.58 \\
  En-Zh & 9.93 \\
  En-Ja & 7.22 \\
  \hline
  \end{tabular}
  }
  \caption{Statistics of WMT bilingual data used in pre-training experiments for multi-target translation.\label{tab:pre-train}}
\end{table}

To generate the pseudo data, taking En$\to$De/Fr as an example, we first train individual Transformer models for En$\to$De and En$\to$Fr using the trilingual data. We then use the En$\to$De model to translate half of the English source $\src_{1/2}$ into German $\hat\trg_{1/2}$ and use the En$\to$Fr model to translate the other half of the English source $\src_{2/2}$ into French $\hat\trg_{2/2}$, thus obtaining the pseudo-trilingual dataset $\{(\src_{1/2}, \hat\trg_{1/2}^1, \trg_{1/2}^2), (\src_{2/2}, \trg_{2/2}^1, \hat\trg_{2/2}^2)\}$ that is as large as the original data. Pseudo datasets for De$\to$En/Fr and En$\to$Zh/Ja are generated similarly.

For Bi-directional translation, we reuse the same trilingual data as described above. Pseudo data is generated by first training individual En$\to$De$_{L2R}$ and En$\to$De$_{R2L}$ Transformer models; we then follow the same procedure as above. The En$\to$De$_{R2L}$ system is trained on En-De trilingual data with the German reference simply inverted. Pseudo data for the other language pairs is generated similarly.

\begin{table*}[!t]
  \center
  \scalebox{0.8}{
  \begin{tabular}{l|ccc|ccc|ccc|l|l}
  \hline
  Model  & De-En & De-Fr & SIM & En-De & En-Fr & SIM & En-Zh & En-Ja & SIM & Avg BLEU & Avg SIM\\
  \hline
  \texttt{base}      & 33.0 & 26.2 & 90.09 & 28.8 & 38.1 & 91.77 & 28.4 & 13.7 & 81.99 & 28.0        & 87.95 \\
  \texttt{multi} & 31.6 & 26.0 & 92.24 & 27.8 & 35.7 & 91.99 & 31.5 & 11.6 & 83.60 & 27.4 (-0.6) & 89.28 (+1.33) \\
  \hline
  \texttt{indep}  & 34.3 & 26.8 & 90.78 & 29.8 & 38.7 & 92.00 & 29.1 & 14.3 & 83.22 & 28.8 (+0.8) & 88.67 (+0.72) \\
  \texttt{dual}   & 32.0 & 22.1 & 91.13 & 29.4 & 37.2 & 91.75 & 28.9 & 14.2 & 84.24 & 27.3 (-0.7) & 89.04 (+1.09) \\
  \hline
  \texttt{indep} ps   & 33.5 & 26.7 & 91.05 & 29.1 & 38.7 & 92.52 & 28.8 & 13.8 & 83.37 & 28.4 (+0.4) & 88.98 (+1.03) \\
  \texttt{dual} ps    & 33.0 & 26.5 & 91.55 & 29.3 & 37.9 & 92.49 & 28.7 & 14.3 & 83.80 & 28.3 (+0.3) & 89.28 (+1.33)\\
  \hline
  \texttt{indep} FT  & 37.0 & 29.5 & 91.98 & 32.1 & 42.0 & 92.53 & 32.0 & 16.5 & 84.74 & 31.5 (+3.5) & 89.75 (+1.80) \\
  \texttt{dual} FT   & 36.8 & 28.4 & 91.87 & 31.8 & 41.0 & 92.80 & 32.6 & 16.5 & 85.01 & 31.2 (+3.2) & 89.89 (+1.94) \\
  \hline
  \texttt{indep} FT+ps & 36.4 & 29.1 & 92.47 & 31.4 & 40.9 & 92.97 & 31.8 & 16.0 & 84.64 & 30.9 (+2.9) & 90.03 (+2.08) \\
  \texttt{dual} FT+ps  & 36.3 & 29.2 & 92.56 & 31.8 & 40.9 & 92.91 & 32.1 & 16.0 & 84.62 & 31.1 (+3.1) & 90.03 (+2.08) \\
  \hline
  \end{tabular}
  }
  \caption{BLEU and similarity scores of multi-target models on \texttt{tst2015}. Similarity scores (SIM) are computed as the cross-lingual similarity between the two target translations. Pseudo (ps) refers to models trained from scratch with synthetic reference data. FT indicates models fine-tuned from the pre-trained multilingual (\texttt{multi}) model. FT+ps refers to models fine-tuned using synthetic reference data.\label{tab:multi-target-2015}}
\end{table*}
\section{Details of Data for Code-switched Input Translation}

We use the same WMT14 En-Fr data as in previous section to generate artificial code-switched sentences. These are obtained by randomly replacing small chunks in one sentence by their translation according to the following procedure. We first compute word alignments between parallel sentences using \texttt{fast\_align}\footnote{\url{https://github.com/clab/fast\_align}} \citep{Dyer13simple} in two directions, then apply a standard symmetrization procedure. Using the algorithm of \citet{Crego05reordered}, we then identify bilingual phrase pairs $(\src,\trg)$ extracted from the symmetrized word alignments under the condition that all alignment links outgoing from words in $\trg$ reach a word in $\src$, and vice-versa.

For each pair of parallel sentence, we first randomly select the primary language;
then the number of substitutions $r$ to perform using an exponential distribution as:
\begin{equation}
P(r=k) = \frac{1}{2^{k+1}} \quad \forall k = 1, \dots, \operatorname{rep},
\label{eq:prob}
\end{equation}
where $\operatorname{rep}$ is the maximum number of replacements. We also make sure that the actual number of replacements never exceed half of either the original source or target sentence length, adjusting the actual number of replacements as:
\begin{equation}
n = \min (\frac{S}{2}, \frac{T}{2}, r),
\end{equation}
where $S$ and $T$ are respectively the length of the source and target sentences. We finally choose uniformly at random $r$ phrase pairs and replace these fragments in the primary language by their counterpart in the secondary language.

A shared vocabulary built with a joint BPE of 32K merge operations is used for CSW source as well as for English and French targets.

\section{Details of Data for Generating Translations of Varying Formalities\label{sec:detail-polite}}

We reuse the data of \citet{Sennrich16controlling}.\footnote{\url{http://data.statmt.org/rsennrich/politeness/}} The training data consists of OpenSubtitles2012 En-De data with $5.58$M sentence pairs, out of which $0.48$M of German reference are annotated as polite and $1.06$M as impolite. The rest is deemed neutral. The annotation tool is based on the ParZu dependency parser\footnote{\url{https://github.com/rsennrich/ParZu}} and an annotation script that is also released with the data. Polite/Impolite tags are based on an automatic analysis of the German side according to rules described in \cite{Sennrich16controlling}. The \texttt{test} set that we use as development set is a random sample of $2000$ sentences from OpenSubtitles2013. We use the \texttt{testyou} set as our main test set, which consists of $2000$ random sentences also extracted from OpenSubtitles2013 where the English source contains a 2nd person pronoun \textit{you(r(s(elf)))}. 

We built shared vocabulary with a joint BPE of 32K merge operations. When fine-tuning the dual decoder models, we also randomly extract an equivalent number of neutral sentences as the polite and impolite ones, i.e.\ $1.54$M. Reference of neutral sentences is thus identical for both polite and impolite targets. The overall fine-tuning data thus comprises $3.07$M sentences.

\section{More Results of \texttt{tst2015} for Multi-target Translation\label{sec:res-appendix}}

Table~\ref{tab:multi-target-2015} reports results for the multi-target translation experiments of Section~\ref{sec:multi-target} using the IWSLT \texttt{tst2015}, a setting that is also used in \cite{He21synchronous}.

\end{document}